\documentclass[runningheads]{llncs}

\usepackage{graphicx}
\usepackage{amsmath,amssymb}
\usepackage{booktabs}
\usepackage{multirow}
\usepackage{array}
\usepackage{algorithm}
\usepackage{subcaption}
\usepackage{algorithmic}
\usepackage[utf8]{inputenc}
\usepackage{url}
\usepackage{xcolor}
\usepackage{enumitem}
\usepackage[hidelinks]{hyperref}
\hypersetup{colorlinks=true,linkcolor=blue,citecolor=blue,urlcolor=blue}
\usepackage{url}
\usepackage{cleveref}

\begin{document}

\title{Axial-UNet: A Neural Weather Model for Precipitation Nowcasting}
\titlerunning{Axial-UNet}



\author{Sumit Mamtani \and
Maitreya Sonawane}
\authorrunning{S. Mamtani and M. Sonawane}
%
\institute{New York University, New York, USA \\
\email{\{sm9669, mss9240\}@nyu.edu}}
\maketitle


\begin{abstract}
   Accurately predicting short-term precipitation is critical for weather-sensitive applications such as disaster management, aviation, and urban planning. Traditional numerical weather prediction can be computationally intensive at high resolution and short lead times. In this work, we propose a lightweight UNet-based encoder–decoder augmented with axial-attention blocks that attend along image rows and columns to capture long-range spatial interactions, while temporal context is provided by conditioning on multiple past radar frames. Our hybrid architecture captures both local and long-range spatio-temporal dependencies from radar image sequences, enabling fixed lead-time precipitation nowcasting with modest compute. Experimental results on a preprocessed subset of the HKO-7 radar dataset demonstrate that our model outperforms ConvLSTM, pix2pix-style cGANs, and a plain UNet in pixel-fidelity metrics, reaching PSNR 47.67 and SSIM 0.9943. We report PSNR/SSIM here; extending evaluation to meteorology-oriented skill measures (e.g., CSI/FSS) is left to future work. The approach is simple, scalable, and effective for resource-constrained, real-time forecasting scenarios.
\end{abstract}
\keywords{Precipitation nowcasting \and UNet \and Axial Attention \and Transformer \and radar \and deep learning}

\section{Introduction }
\label{sec:intro}

Deep neural networks (DNN)\cite{sze2017efficient,2015,barnes2020identifyingopportunitiesskillfulweather,2019,2021} have been successfully applied in many diverse domains such as image classification, video analysis, language modeling and translation, medical imaging, and weather. Recently, there has been increasing interest in using DNNs to generate and improve weather
forecasting, which is an unsupervised representation problem. In these kinds of problems, next-frame prediction is a new, promising direction of research in computer vision, predicting possible future images by presenting historical image information. Recent neural network-based weather models highlight this trend toward learned forecasting~\cite{bodnar2024aurora}.
 \\
Weather forecasting is the prediction of future weather conditions, such as precipitation, temperature, pressure, and wind, and is fundamental to both science and society. Our particular interest is in the area of nowcasting, a term used to describe high-resolution, short-term (e.g., 0 to 2 hours) weather forecasts of precipitation or other meteorological quantities. The precipitation nowcasting field helps in the accurate prediction of rainfall over an area by looking at radar images. The field deals with the generation of the radar image at some points in the near future.\\  
One of the various architectures researchers have tried to implement is Conv-LSTM\cite{shi2015convolutional,staudemeyer2019understanding,patel2018precipitation}, which has shown great results in dealing with time-series data because it is pretty good at extracting patterns in the input feature space, where the input data spans over long sequences. The gated architecture of LSTMs has the ability to manipulate the memory state, making it more subtle for such problems. \\
With the continuous input of data from one end, we know that the prediction needs to be swift while dealing with a huge amount of data. Following this, some recent works have used UNet for this problem and shown improved results for image-to-image translation problems~\cite {choi2019traffic,trebing2021smaatunet}.
\\ 
Recently, conditional GANs (cGANs)~\cite{mirza2014conditional} have become widely popular under the domain of image-to-image translation\cite{isola2018imagetoimage}. As we require a model that would learn from previous inputs in the sequence, we tried to make use of this property, where cGANs predict N future radar frames given M past-conditional frames.\\
In this Paper, we also introduced an encoder-decoder architecture~\cite{mo2021encoderdecoder} with the Axial Transformer~\cite{ho2019axial} in our proposed model, a simple yet effective self-attention-based \cite{vaswani2017attention,yun2021transformer} autoregressive model for data organized as multidimensional tensors. Rather than applying attention to a flattened string of tensor elements, Axial Transformer instead applies attention along a single axis of the tensor without flattening, so this is referred to as “axial attention”. Since the length of any single axis (that is, the height or width of an image) is typically much smaller than the total number of elements, an axial attention operation enjoys a significant saving in computation and memory over standard self-attention. We achieve competitive results on a preprocessed dataset by training the encoder-decoder with the Axial Attention blocks. Our contributions are summarized as follows:
\begin{itemize}[leftmargin=*]
  \item \textbf{Efficient encoder–decoder for fixed lead-time nowcasting.} We propose a lightweight encoder–decoder that ingests $M$ past radar frames and predicts the next frame at a fixed lead time, maintaining native resolution while keeping compute modest.
  \item \textbf{Axial attention for long-range spatial context.} We insert axial-attention blocks that attend along rows and columns, capturing large-scale advection patterns with near-linear memory; temporal context is supplied by the multi-frame encoder.
  \item \textbf{Controlled evaluation on an HKO-7 subset.}
All models share the same preprocessing (grayscale $128\times128$) and the same train/val/test split. Each baseline uses its typical \emph{number of input frames}---ConvLSTM (15), cGAN (4), and UNet/ours (16)---and we evaluate fixed lead-time (next-frame; $M\!\to\!1$) predictions, extending to longer horizons via autoregression.
  \item \textbf{Quality gains.} The axial-attention variant improves PSNR/SSIM over strong baselines(see \autoref{tab:quant}).
\end{itemize}

\section{Related Work}
\label{sec:formatting}

Previously, the Conv-LSTM model has shown promising results on the next frame prediction problem on the Moving MNIST dataset \cite{shi2015convolutional}.
The dataset was initially created in the context of Unsupervised Learning of Video Representations \cite{srivastava2016unsupervised} and used LSTM to learn the representation of the video sequence. As the radar images are also a kind of time series data, and we know that clouds can not abruptly change direction or disappear, we know that there are some motion parameters associated with a continuous sequence of radar images, too. Using the same idea of next frame prediction in a video sequence, ConvLSTM can also be applied to a dataset of radar images. \\
UNet is one of the recent models that has shown its ability to predict the next frame of time-series data very well \cite{choi2019traffic}. In the paper, as part of the Traffic4cast challenge 2019, UNet was used to predict short-term traffic flow volume. The input and output of the model were the same sizes, that are also relevant in our context, as we need to reproduce radar images of the same location sometime in the future. Another paper that actually implemented UNet for the weather forecasting problem \cite{trebing2021smaatunet} saw a significant increase in results on the dataset consisting of precipitation maps from a region of the Netherlands and a binary image of cloud coverage of France. The size of the model here is very small compared to previous models that were used to solve the same problem, which is also a significant advantage, considering the latency requirement of our problem statement.\\
 Researchers have now tried an observations-driven approach for probabilistic nowcasting using deep generative models (DGMs). DGMs are statistical models that learn probability distributions of data and allow for easy generation of samples from their learned distributions\cite{ruthotto2021introduction,2021}. As generative models are fundamentally probabilistic, they can simulate many samples from the conditional distribution of future radar given historical radar.\\
One category of DGM model is GANs \cite{goodfellow2014generative}. GANs learn a loss that tries to classify if the output image is real or fake, while simultaneously training a generative model to minimize this loss. Blurry images will not be tolerated since they look obviously fake. Because GANs learn a loss that adapts to the data, they can be applied to a multitude of tasks that traditionally would require very different kinds of loss functions.
\paragraph{\textbf{Summary and gap}.}
ConvLSTM- and GAN-based approaches capture temporal dynamics but struggle with long-range spatial dependencies or require heavy memory. Plain UNet models are efficient but lack explicit mechanisms for large-scale advection. This motivates our design: a lightweight encoder–decoder augmented with \emph{axial attention} to model row/column interactions efficiently.
\\
\section{Method: UNet with Axial Attention (Axial-UNet)}
\label{sec:method}
\noindent
Given $M$ past radar frames (here $M{=}16$) sampled every 6 minutes, our model predicts the next frame at a fixed lead time; longer horizons are obtained via autoregression. The backbone is a UNet-style encoder--decoder, and we insert \emph{axial-attention} blocks that attend along rows and columns to capture large-scale advection with modest memory overhead.

\subsection{Backbone encoder--decoder}
UNets, first introduced in the Convolutional Networks for Biomedical Image Segmentation paper\cite{ronneberger2015unet}, have been able to expand their use case from image segmentation to predicting the future sequence too. The architecture consists of a contracting path to capture context and a symmetric expanding path that enables precise localization. Encoder (downsampling path) extracts a meaningful feature map from an input image. As is standard practice for a CNN, the Encoder doubles the number of channels at every step and halves the spatial dimension. Next, the Decoder (upsampling path) actually upsamples the feature maps, where at every step, it doubles the spatial dimension and halves the number of channels (opposite to what an Encoder does). \\
The contractive path (Encoder) consists of the repeated application of two 3x3 convolutions (padding=1), each followed by a rectified linear unit (ReLU) and a 2x2 max pooling operation with stride 2 for downsampling. At each downsampling step, we double the number of feature channels. As the image size we use (128 x 128) is very small to be downsampled too much, we couldn't traverse the contractive path as much as given in the reference papers, i.e., rather than having out channels as 1024 in the final output of our Encoder, we had 256 as our number of out channels. Hence, the sequence of in and out channels for each block in the Encoder consisted of [16,64,128,256].\\
The decoder layer is explained as follows: every step in the expansive path consists of an upsampling of the feature map followed by a 2x2 convolution (up-convolution) that halves the number of feature channels, a concatenation with the correspondingly cropped feature map from the contracting path, and two 3x3 convolutions, each followed by a ReLU. The cropping is necessary due to the loss of border pixels in every convolution. The \textit{ConvTranspose2d} operation performs the up-convolution, and again the same block consisting of Conv2D and ReLU in between is used to half the number of channels. The output of this Decoder is a 128 x 128 image for each batch, which can again be compared to the target and initiate the learning of our model.
\\
The usage of residual skip connections helps alleviate the vanishing gradient problem, allowing for UNet models with deeper neural networks to be designed. Each residual unit can be
denoted by the following expressions:

\begin{equation}
 y_{\ell} = h(x_{\ell}) + F(x_{\ell}, W_{\ell}),\quad x_{\ell+1}=y_{\ell}.
\end{equation}

To handle the time series data that we have, we could have probably used Conv-3D and MaxPool-3D layers in our UNet. But as the data was preprocessed into a GRAYSCALE image, we could just substitute the number of frames as the number of input channels, an idea inspired by SmaAt-UNet. \cite{trebing2021smaatunet}. Recent UNet variants with attention also show benefits in remote sensing nowcasting~\cite{zhang2025raunet}.
\\

\subsection{Axial attention blocks}
Our proposed model is based on axial attention, a simple generalization of self-attention that naturally aligns with the multiple dimensions of the tensors in both the encoding and the decoding settings. As we know, Attention mechanisms have become an integral part of compelling sequence modeling and transduction models in various tasks, allowing modeling of dependencies without regard to their distance in the input or output sequences. So we have used Axial Transformers, an axial-attention-based autoregressive model for images and other data organized as high-dimensional tensors.\\
The sequence of images, which are the first 16 frames from the sequence of length 20, is fed into our Downsampler, which is the same as the Encoder of our UNet model used before. This acts as an image processing pipeline and the first stage of our model. The next layer is a Decoder layer that outputs a representation of the time series, i.e., integrating the information over time. The Internal layers of both Encoder and Decoder are made of several layers of CNN. We produce one feature map per input image from a sequence of length 16. Previously, in the simple UNet model, we compressed all information within a single image output from the model, but now these 16 images will act as input to our next stage, the Axial Transformer \cite{ho2019axial}. We adopt axial attention to model long-range spatial interactions with reduced memory cost in radar imagery~\cite{xie2024axialecho}.
\\
Axial attention can be straightforwardly used within standard Transformer layers to produce Axial Transformer layers. The theoretical foundation for axial attention is established in prior work \cite{ho2019axial}, which provides the mathematical formulation and proof for decomposing two-dimensional self-attention into separable row and column-wise operations. In this study, we adopt their proven formulation within our encoder-decoder framework, demonstrating its empirical stability and effectiveness for spatio-temporal precipitation nowcasting. The basic building blocks are the same as those found in the standard Transformer architecture: \\

\textbf{Inner Decoder}: using masked row attention layers to create a “row-wise” model:
\begin{algorithm}

    h ← Embed(x)\\
    h ← ShiftRight(h) + PositionEmbeddings\\
    $h \leftarrow \text{MaskTransformerBlock2}(h) \times L_{\text{row}}$
    
\end{algorithm}
$L_{row}$ is the number of masked row attention blocks applied to h. The operation ShiftRight shifts the input right by one pixel.
PositionEmbeddings is a tensor of position embeddings that inform the attention layers of the position. x is the gray scale frame.\\

\textbf{Outer Decoder}: Each pixel in the model depends on previous pixels in its own row. To capture all previous rows, we insert unmasked row and masked column layers at the beginning of the model as described below:

\begin{algorithm}
    h ← Embed(x)\\
    u ← h + PositionEmbeddings\\
    u ← MaskTransformerBlock1(Block2(u)) × $L_{upper}/2$\\
    h ← ShiftDown(u) + ShiftRight(h) + PositionEmbeddings\\
    $h \leftarrow \text{MaskedTransformerBlock2}(h) \times L_{\text{row}}$
\end{algorithm}

The tensor u represents context captured above the current pixel.  It is computed by unmasked row and masked column attention layers, repeated to a total of $L_{upper}$ layers.\\
The idea of using a transformer with an encoder-decoder model was inspired by a development in the field of Precipitation Nowcasting at Google, which introduced a Neural Weather model called MetNet \cite{metnet}.\\
Now, what Axial Transformer will try to do is encode information in these feature maps from the space around each point of interest. Hence, attending to parts that are relevant to justify the motion of a cloud patch in a radar image. The way the Transformer works is - we have a series of feature maps that are continuous with respect to time. Each pixel from the latest frame emits a query vector, and each of the pixels from the older image feature maps emits a key. And each of the pixel-emitting queries can look at (attend) each of the pixels in the lower layer (keys). Hence, we can incorporate long-range dependencies by aggregating information from the downstream and increasing the resolution. But in Axial Transformers, we will specifically attend to the pixels that are either in the same row or the same column in the images produced. This saves the required memory for the computation over the layers of attention while aggregating information over the spatial dimensions. The code implementation of the Transformer was inspired by the PyTorch implementation of axial attention available on GitHub \footnote{\url{https://github.com/lucidrains/axial-attention}}.\\
The final output of this model is a distribution across 128 frames, output from the transformer head, out of which we need to take the mean value as the predicted state of a pixel after the prediction. This predicted tensor, flattened, along with the target image tensor, also flattened, is sent to calculate the loss and start model learning.

\subsection{Baselines}
We compare against (i) ConvLSTM and (ii) a pix2pix-style conditional GAN (cGAN), in addition to a plain UNet backbone. All baselines share our preprocessing and split (grayscale $128{\times}128$). Input lengths follow customary practice---ConvLSTM (15), cGAN (4), UNet/ours (16)---and we evaluate fixed lead-time ($M\!\to\!1$) predictions, extending to multiple frames via autoregression.

\subsubsection{ConvLSTM}
We use ConvLSTM as a baseline, following Shi et al.~\cite{shi2015convolutional} on the HKO-7 dataset. We adapt a PyTorch implementation\footnote{\url{https://github.com/sladewinter/ConvLSTM}} originally developed for Moving MNIST to our radar sequences. ConvLSTM extends fully connected LSTMs by replacing affine transforms with convolutions in both the input-to-state and state-to-state transitions, enabling end-to-end sequence-to-sequence modeling for precipitation nowcasting. 

Our implementation uses a three-layer encoder–decoder with 64 \(3\times3\) kernels per layer and padding 1 to preserve spatial size, ReLU activations, and the Adam optimizer. Each ConvLSTM cell comprises a Conv2D layer and input, output, and forget gates with sigmoid activations; the gates control memory updates and retention of past state. In our setup we condition on \(M{=}15\) input frames and predict the next frame, rolling out longer horizons autoregressively. For a recent survey of deep learning for precipitation nowcasting and datasets/metrics, see~\cite{an2024survey}.
\\

\subsubsection{cGANs}
GANs are generative models that learn a mapping from a random noise vector z to an output image y, G: z → y. But conditional GANs learn a mapping from observed image x and random noise vector z, to y, G: {x, z} → y.
The generator G is trained such that it produces outputs that cannot be
distinguished from “real” images by a trained discriminator (D), which is trained to do as well as possible at detecting the generator’s fake images.\\
The objective of a conditional GAN can be expressed as:\\
\begin{align}
\begin{split}
    L_{cGAN}( G,D) =E_{x,y}[ log( D( x,y)] \ +\\ E_{x,z}[ log( 1-D(  x,G( x,z))]
\end{split}
\end{align}

where G tries to minimize this objective against an adversarial D that tries to maximize it, so the overall objective function is given by:-\\
\begin{align}
\begin{split}
    G^{*} =\ arg\ min_{G} (\ arg\ max_{D} (L_{cGAN}( G,D)))
\end{split}
\end{align}
\begin{figure}
  \centering
  \includegraphics[width=\linewidth]{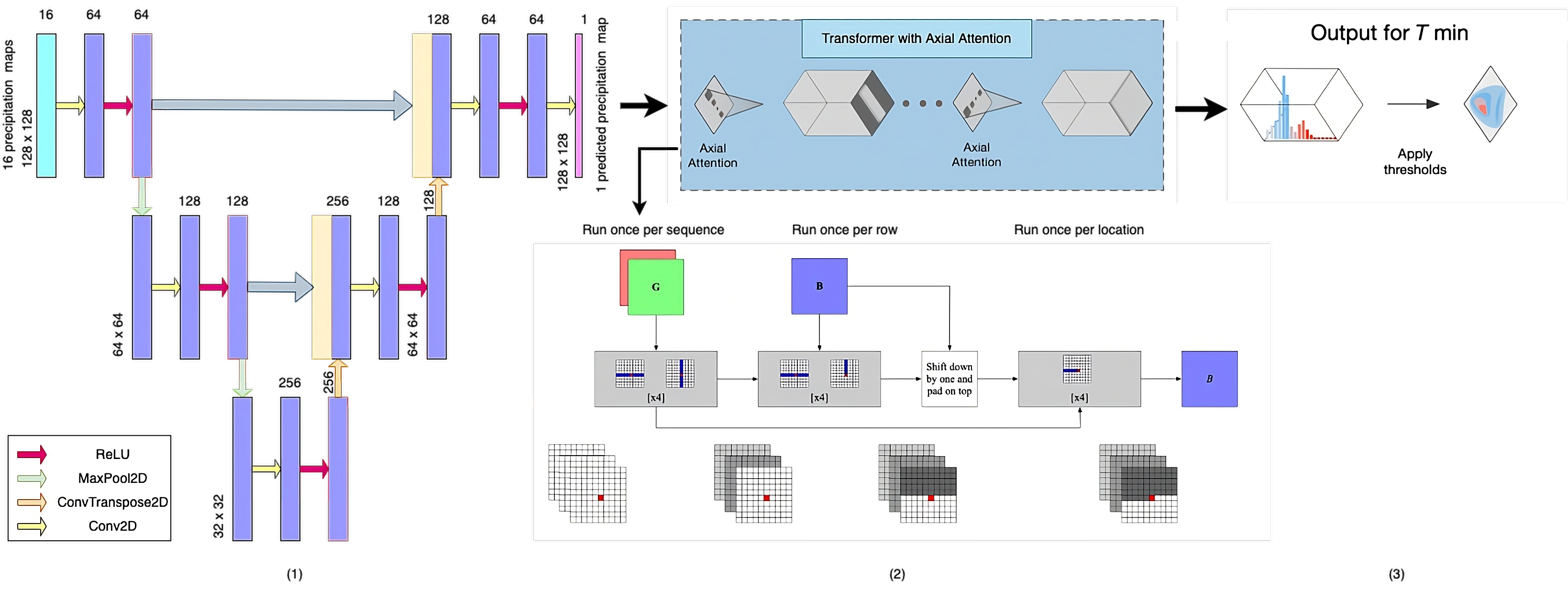}
  \caption{Architecture of Axial-UNet (UNet with axial attention). Part (1) is our UNet model that encompasses an encoder and decoder using Conv2D. Part (2) is the transformer \cite{metnet} that attends every row and column for each location (pixel), hence an axial attention \cite{ho2019axial}. Part (3) is the distribution we get as our final output, of which we take the mean to make a final prediction for T$^{th}$ minute}
  \label{fig:UNet with Axial Transformer model}
\end{figure}
$L(G,D)$ is known as adversarial loss. We have also used Pixelwise L1 content loss to train the generator and discriminator model. So the generator task is to not only fool the discriminator but also to be near the ground truth output in an L1 sense. So our total loss comprises the addition of both the pixelwise loss and adversarial loss. We have used L1 distance rather than L2 as L1 encourages less blurring. So L1 loss are as follows is given by:\\
\begin{align}
\begin{split}
    L_{L1}( G) \ =\ E_{x,y}[ \ ||\ y\ -\ G( x,z) \ ||_{1}]
\end{split}
\end{align}
So final combined loss function is given by:\\
\begin{align}
\begin{split}
    G^{*} =\ arg\ min_{G} \ (arg\ max_{D} (L_{cGAN}( G,D) \ +\\ \lambda L_{L1}( G)))
\end{split}
\end{align}
where $\lambda $ is a hyper-parameter, and in our implementation, we have set its value to 100.
\subsubsection{Generator}
The generator consists of a series of layers that progressively downsample until a bottleneck layer is reached, after that we reverse the same process, i.e, we upsample. Such a network requires that all information flow passes through all the layers, including the bottleneck. So there is a low level of information shared, like the prominent edge locations between input and output, which produces higher-quality images than a simple CNN generator model.\\
We have added skip connections to avoid the vanishing gradient problem and to incorporate information from earlier layers. This is the general shape of a UNet architecture. Specifically, we add skip connections between each layer $i$ and layer $n - i$, where $n$ is the total number of layers. Each skip connection simply concatenates all channels at layer $i$ with those
at layer $n - i$.
\subsubsection{Discriminator}
The discriminator is a type of Convolution Neural Network(CNN) model. The discriminator has 4 sequential blocks having 64, 128, 256, 512 filters respectively. Each sequential block consists of the convolutional layer, batch-normalization layer, and activation layer as LeakyReLU. The generator tries to fool the discriminator by creating fake images. The discriminator is trained in such a way that it tries to distinguish real data from the data created by the generator.
\section{Experimental Results}
In this section, we first introduce related datasets and implementation details in our experiments. Then, we report results and a comparison of all the experiments conducted on each of the models used.
\subsection{Datasets}
The dataset used in this project is a part of dataset used in paper ConvLSTM for Precipitation Nowcasting \cite{shi2015convolutional}. The dataset for the original research work consisted of HKO-7 dataset, which
contains radar echo data from 2009 to 2015 near Hong Kong. Along with the radar images, the dataset also contains proportions of rainfall events with different rain-rate thresholds. \\
The HKO-7 dataset used in the baseline research contains radar CAPPI reflectivity images, which have a resolution of 480×480 pixels, are taken from an altitude of 2km, and cover a 512km × 512km area centered in Hong Kong. The data are recorded every 6 minutes, and hence there are 240 frames per day. The pixel values are clipped between 0 and 255. There are some noisy radar images in the dataset generated by factors like ground clutter, sea clutter, anomalous propagation, and electromagnetic interference. \\
The complete dataset with all the CSV and pkl files was not available publicly. But a part of it, i.e., the radar images that can be conveniently used in our problem, are available publicly \cite{DVN/2GKMQJ_2019}. We have continuous radar images from 514 folders. Each folder contains a variable number of radar images sampled, but all of them are continuous. As the radar echo maps arrive in a stream, nowcasting algorithms can apply online learning to adapt to the newly emerging spatiotemporal patterns. \\
This was the most convenient dataset we could use, because the datasets used in previous research under this topic were too large to be handled on the GPU provided (datasets of size in TBs). Other datasets were either paid or not publicly available. Even the dataset that we have used, is a part of an original dataset. But it was enough to carry out several experiments on various models used. 
\subsection{Implementation Details}

After taking the dataset as our input, first, we had to ensure that the number of radar images in each folder's sequence remains the same. This is due to the fact that we will be sending the dataset in batches for training and testing, and we need to have a uniform number of images in each batch. Hence, we clipped the dataset to 240 images per folder (if the number of radar images in a folder is greater than 240, if it is less than 240, we reject that folder), i.e., as soon as the number of images in that particular folder goes past 240, we stop taking that folder's images as input. We found out that the dataset now had 482 sequences of length 240. While we input each image, we also convert the image into grayscale, as we need to map the cloud motion in the radar images, and having RGB channels in the images will only cost computation time and power without much significant difference in the results.\\
At first, we tried to send this data directly for training in batches. But, due to computational power limitations, we could not send a batch of size 240 for training. The sequence of images had to be broken to the length of 20, i.e., instead of 482 sequences of length 240, now we had 5784 sequences of length 20. Now sequence of this length can be comfortably loaded for training and testing. \\
Now that we had a sequence of data, we also needed to filter out the images that were either too dark or too light. There were some faulty images in our dataset, as previously talked about, because of faulty radar images. So we first stored the summation of all the pixel values in each image of our dataset in a list. This is important to normalise our dataset, as then we found the 25$^{th}$ and 75$^{th}$ percentile of the list values. These give us the range we will use to filter images whose sum of pixel values is above or below the specified range. \\
Now, going through each sequence, we again calculated the sum of pixel values of each image in 5784 sequences. If the sum is outside the range, we count that as a bad frame. If the number of bad frames in a sequence exceeds 10, we discard that sequence as a whole, as we can't discard a few frames from a sequence; discontinuity in frames would make them useless. Now, once the data is cleaned, we are left with 2901 sequences of length 20, each being an image of size 128 x 128. The sequences were then loaded in an .npy file to be used later. As observed, the usable images left at last were almost half of the dataset we created at the start, and this was one of the reasons why we couldn't use a higher batch size while training our models.\\
\begin{table*}[t!]
\centering
\renewcommand{\arraystretch}{1.2} 
\newcolumntype{C}{>{\centering\arraybackslash}m{0.18\textwidth}} 

\begin{tabular}{C C C C C}
\rotatebox[origin=c]{90}{\textbf{Target}} &
\includegraphics[width=\linewidth]{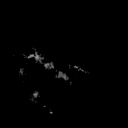} &
\includegraphics[width=\linewidth]{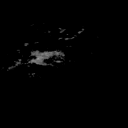} &
\includegraphics[width=\linewidth]{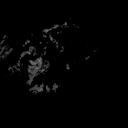} &
\includegraphics[width=\linewidth]{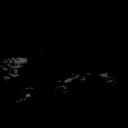} \\

\rotatebox[origin=c]{90}{\textbf{Output}} &
\includegraphics[width=\linewidth]{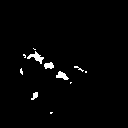} &
\includegraphics[width=\linewidth]{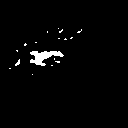} &
\includegraphics[width=\linewidth]{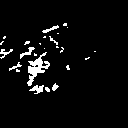} &
\includegraphics[width=\linewidth]{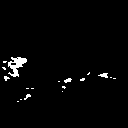} \\

\textbf{Configuration} & 
\textbf{ConvLSTM} & 
\textbf{cGANs} & 
\textbf{UNet} & 
\textbf{Encoder-Decoder with Axial Transformer}
\end{tabular}

\caption{\label{tab:result_comparison}While testing each model, we first input a sequence of images ($t_0, t_1, ..., t_m$) into the model and obtain one image ($p_1$) as output. This predicted image $p_1$ is then used with the next sequence ($t_1, ..., t_m$) to predict the next radar image, and so on. The top row shows the target images, and the second row shows the outputs. The proposed model achieves the highest quality of prediction.}
\end{table*}

Once the .npy file is loaded, we randomly shuffle these sequences without altering their internal continuity in frames, for uniform distribution of types of images for training, validation, and testing. Out of these, 2000 sequences were used in training dataset, 500 for validation, and 400 for testing purpose. Whenever the numpy array representing the images in the dataset is loaded, the batch is normalised by dividing by 255, to bring each pixel value between 0 and 1. Shuffle is kept true to make the distribution among the types of images uniform. \\
The sequence length given to train Conv-LSTM was 15, so that it can count next 5 frames as target and make a prediction on those. For cGANs, the input sequence length was 4, as it was an image-to-image translation technique and produced one image per input image, hence predicting 4 frames from 4 inputs. For UNet and UNet with axial attention, we had to give 16 frames as input to the model, as we needed to ensure downsampling and later upsampling produce images (given in number of channels dimensions) of the same number at each corresponding layer. The batch size for ConvLSTM = 2, cGANs = 4, UNet = 4, Axial-UNet = 1. Every model was given a batch size with limitations due to the risk of the GPU running out of memory or model overfitting too soon. \\
For a given input sequence, 1 frame was predicted as output from every model. Now the same frame was used as input while predicting the next frame and so on. Hence, by this method, we were able to predict 4--5 frames in the future using the original sequence of the input image.\\
Number of epochs used to train the models is variable due to the effect of overfitting for larger and deeper models. ConvLSTM: 5, cGANs: 10, UNet: 5, Axial-UNet: 15. Optimizer used is Adam, with a learning rate of: \textit{1e-3} for ConvLSTM, \textit{1e-4} for cGANs, \textit{1e-3} for UNet, \textit{1e-4} for Axial-UNet. The loss criterion for each of the models used is \textit{MSELoss}. \\
Given a noise-free flattened monochrome image I and its flattened noisy approximation Image K, the \textit{MSELoss} between two tensors is defined as:
\begin{equation}
    MSELoss\ =\ \frac{1}{2}\sum _{i=0}^{n-1}( \ I_{i} \ -\ K_{i} \ )
\end{equation}

The metrics we used to compare the results of different models were PSNR and SSIM. PSNR: Peak signal-to-noise ratio is the ratio between the maximum possible power of a signal and the power of corrupting noise that affects the fidelity of its representation. \\
\begin{equation}
    PSNR\ =\ -10\ \log_{10}( MSE)
\end{equation}
MSE is defined as the mean squared error between a noise-free, flattened monochrome image and its flattened noisy approximation, Image K\\
SSIM: Structural Similarity Index (SSIM) measures the similarity between two images by quantifying the image degradation of one image with respect to another. \\
The SSIM between two images x and y of common size N×N is:\\
\begin{equation}
    SSIM( \ x,y) \ =\ \frac{( 2u_{x} u_{y} +c_{1})( 2\sigma _{xy} +c_{2})}{\left( u_{x}^{2} +u_{y}^{2} +c_{1}\right)\left( \sigma _{x}^{2} +\sigma _{y}^{2} +c_{2}\right)}
\end{equation}
where $\mu _{x}$ is the average of x, $\mu _{y}$ is the average of y, $\sigma _{x}^{2}$ is the variance of x; $\sigma _{y}^{2}$ is the variance of y and $c_{1}$ and $c_{2}$ are constants.

\subsection{Results and Comparisons}

We limit evaluation to PSNR/SSIM. Event/object-based skill and cell-tracking metrics are standard in meteorology~\cite{ritvanen2025celltracking}; comparison to recent diffusion/transformer models is left for future work.
While training each of the model, we calculated the loss for training and cross-validation data, the results of which are given below:

\begin{figure*}[htbp]
    \centering
    
    \begin{subfigure}[b]{0.45\textwidth}
        \centering
        \includegraphics[width=\linewidth]{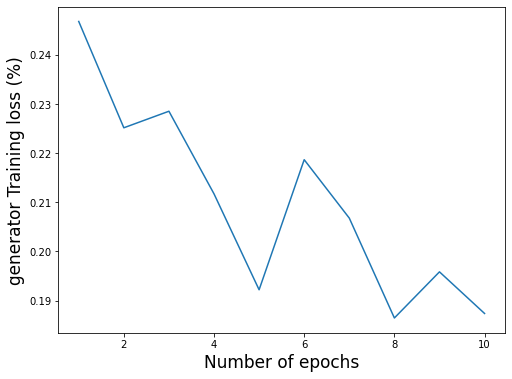}
        \caption{cGANs training loss}
    \end{subfigure}
    \hfill
    \begin{subfigure}[b]{0.45\textwidth}
        \centering
        \includegraphics[width=\linewidth]{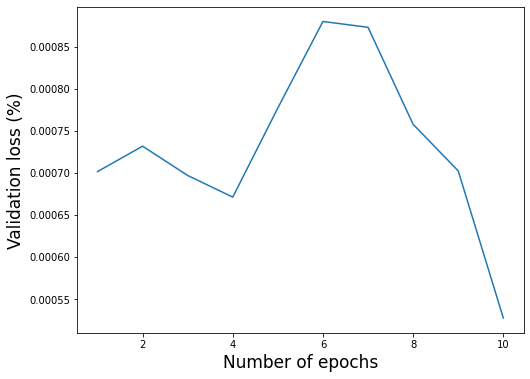}
        \caption{cGANs validation loss}
    \end{subfigure}
    
    \begin{subfigure}[b]{0.45\textwidth}
        \centering
        \includegraphics[width=\linewidth]{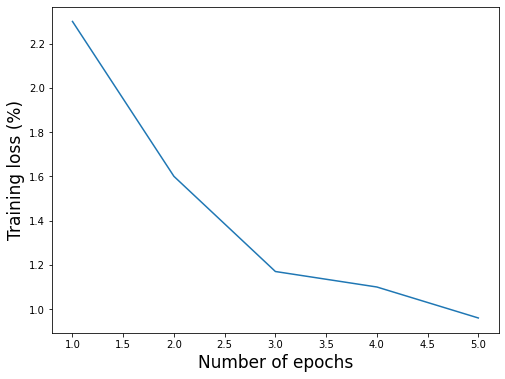}
        \caption{UNet training loss}
    \end{subfigure}
    \hfill
    \begin{subfigure}[b]{0.45\textwidth}
        \centering
        \includegraphics[width=\linewidth]{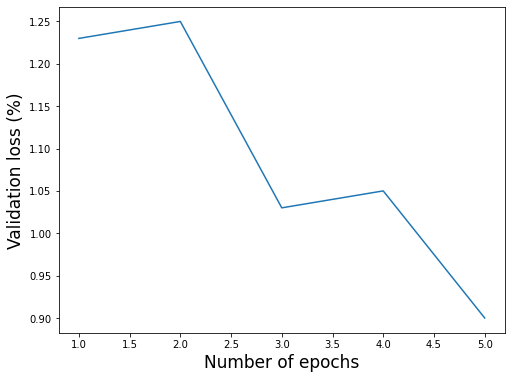}
        \caption{UNet validation loss}
    \end{subfigure}
    
    \begin{subfigure}[b]{0.45\textwidth}
        \centering
        \includegraphics[width=\linewidth]{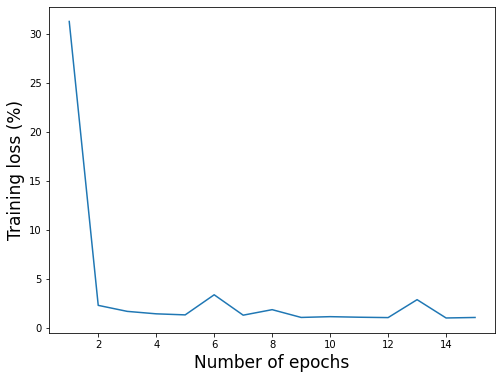}
        \caption{UNet+Transformer training loss}
    \end{subfigure}
    \hfill
    \begin{subfigure}[b]{0.45\textwidth}
        \centering
        \includegraphics[width=\linewidth]{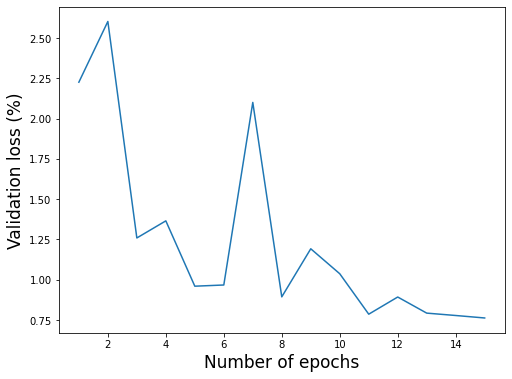}
        \caption{UNet+Transformer validation loss}
    \end{subfigure}
    
    \caption{Training and validation loss curves for (a,b) cGANs, (c,d) UNet, and (e,f) Axial-UNet.}
    \label{fig:training_curves}
\end{figure*}

We see some unusual spikes in the graph, which might be due to faulty radar images and abrupt shuffling of data. \\
The models used in the experiments showed a gradual increase in both metrics, and hence show that our proposed architecture performs better than the baseline model used.
\begin{table}[h!]
\centering
\caption{Comparison on the HKO-7 subset. Higher is better.}
\label{tab:quant}
\label{tab:psnr_ssim}
\renewcommand{\arraystretch}{1.2} 
\begin{tabular}{lcc}
\hline
\textbf{Configuration} & \textbf{PSNR$\uparrow$} & \textbf{SSIM$\uparrow$} \\ \hline
ConvLSTM               & 40.8852       & 0.9710        \\
cGANs                  & 41.1319       & 0.9826        \\
UNet                   & 47.2862       & 0.9929        \\
Encoder-Decoder with Axial Transformer & \textbf{47.6678} & \textbf{0.9943} \\
\hline
\end{tabular}
\end{table}

While the ConvLSTM baseline performed reasonably on our split, a pix2pix-style cGAN improved over it. Treating the task as image-to-image translation (one output frame per input frame) with an adversarial loss helps suppress some noise and sharpen texture compared to pure regression.

The cGAN’s generator is a UNet. Interestingly, the same UNet used as a standalone deterministic model performed even better, likely because our dataset is relatively small and the supervised loss dominates; simpler models can generalize well in this regime. In addition, pix2pix yields fewer distinct training pairs than sequence-to-sequence setups, since each training example maps a single input image to a single target image.

Skip connections in UNet help alleviate vanishing gradients and promote feature reuse, enabling accurate next-frame predictions with a compact backbone. Adding \emph{axial-attention} blocks on top of UNet further improves quality by attending along rows and columns to capture large-scale advection while keeping memory modest.

For evaluation, the model outputs continuous predictions in \([0,1]\) (due to input normalization), and PSNR/SSIM are computed on these continuous fields. Thresholded black-and-white visualizations (e.g., applying a 0.2 cutoff and rescaling to 0–255) are used only for display, not for metric computation.

\section{Discussion}
We studied short–lead-time radar nowcasting with a lightweight encoder–decoder augmented by axial attention. Given $M$ past frames (here $M{=}16$) sampled every 6 minutes, the model predicts the next frame and extends to 4–5 future frames via autoregression. On our HKO-7 subset, the axial-attention variant improves pixel-fidelity metrics (PSNR/SSIM) over ConvLSTM, pix2pix-style cGANs, and a plain UNet. Axial attention preserves coherent structures and sharper boundaries aligned with large-scale advection while keeping memory overhead modest by attending along rows and columns instead of full 2D self-attention.

This study has practical constraints: evaluation is limited to PSNR/SSIM (image fidelity) rather than meteorological skill; a fuller assessment would include event/object-based metrics and lead-time breakdowns. The data are a cleaned, grayscale subset of HKO-7 at $128{\times}128$, which may bias case difficulty; we do not convert reflectivity to rain rate nor assess calibration/uncertainty (we take the mean of the predictive distribution). Baselines follow customary input lengths—ConvLSTM (15), cGAN (4), UNet/ours (16)—and their usual training settings, so the comparison is pragmatic rather than strictly matched-budget.

\section{Future Work}
We plan to extend the study along four axes:
\begin{itemize}[leftmargin=*]
  \item \textbf{Continuous targets and rain rate.} Train on reflectivity in dBZ and predict continuous fields; convert to rain rate via an appropriate $Z$--$R$ relation, and evaluate thresholded rain events rather than binarized visualizations.
  \item \textbf{Meteorology-grade evaluation.} Beyond PSNR/SSIM, report CSI, POD, FAR, FSS, and cell-tracking growth/decay across lead times (e.g., 6--30\,min).
  \item \textbf{Probabilistic outputs.} Retain the distributional head, assess calibration (e.g., CRPS, reliability diagrams), and sample ensembles for uncertainty-aware nowcasts.
  \item \textbf{Data \& baselines.} Use storm- or date-based splits to avoid leakage; quantify cleaning/clutter effects; add persistence/optical-flow controls and stronger deep baselines under the same split; explore higher resolution and multi-channel inputs (multi-elevation radar, satellite bands).
\end{itemize}

{\small
\bibliographystyle{splncs04}
\bibliography{latex/egbib}
}

\end{document}